\PassOptionsToPackage{table}{xcolor}
\documentclass[11pt]{article}
\usepackage[preprint]{acl}

\usepackage{times}
\usepackage{latexsym}
\usepackage[T1]{fontenc}
\usepackage[utf8]{inputenc}
\usepackage{microtype}
\usepackage{graphicx}
\usepackage{booktabs}
\usepackage{multirow}
\usepackage{array}
\usepackage{xcolor}
\usepackage{makecell}
\usepackage{tabularx}
\usepackage{subcaption}
\usepackage{float}
\usepackage{placeins}
\usepackage{needspace}
\usepackage{amsmath}
\usepackage{amssymb}
\usepackage{xspace}
\usepackage{ragged2e}

\makeatletter
\renewcommand\paragraph{\@startsection{paragraph}{4}{\z@}{1.0ex plus .2ex minus .2ex}{-0.8em}{\normalfont\normalsize\bfseries}}
\makeatother

\newcommand{\modelname}{DAFS\xspace}
\newcommand{\up}[1]{\textsuperscript{\textcolor{red!85!black}{\textbf{$\uparrow$#1}}}}

\title{Efficient Frame Selection for Long Videos at Test Time with Attention-Based MLLM Selectors}

\author{
Yilin Wang\textsuperscript{1}\thanks{\ Equal contribution.}\quad
Xiangxi Zheng\textsuperscript{2}\footnotemark[1]\quad
Dongxing Mao\textsuperscript{3}\quad
Linjie Li\textsuperscript{4}\quad
Zhengyuan Yang\textsuperscript{4}\\
\bfseries
Ping Yu\textsuperscript{2}\quad
Rui Yan\textsuperscript{5}\quad
Yuan Yao\textsuperscript{2}\quad
Alex Jinpeng Wang\textsuperscript{3}\\[0.45em]
\normalfont\normalsize
\textsuperscript{1}ZJU\quad
\textsuperscript{2}NJU\quad
\textsuperscript{3}CSU\quad
\textsuperscript{4}Microsoft\quad
\textsuperscript{5}NJUST
}

\begin{document}
\maketitle

\begin{abstract}
Understanding long videos with multimodal large language models (MLLMs) requires selecting a compact set of frames from thousands of candidates, yet identifying the right frames seemingly requires understanding the video first.
We resolve this circular dependency with a simple observation: cross-modal attention at validation-selected extraction layers in MLLMs already provides query-relevant frame evidence without requiring autoregressive generation.
We exploit this property to build \modelname (Dynamic Attention-based Budget-aware Frame Selection), a training-free frame selector.
A lightweight MLLM selector, even with only 2B parameters, can extract frame-level evidence by converting selected-layer attention into relevance scores through query-conditioned aggregation.
This enables cross-frame comparison without autoregressive decoding.
To handle the selector's own context constraint, we formulate the joint allocation of candidate pool size and per-frame token budget as a discrete optimization problem solved by dynamic programming.
Under a 32-frame budget, our selector improves over uniform sampling by up to 6.4 points on Video-MME and outperforms prior training-based selectors under matched frame budgets, while generalizing across selector and answerer backbones, and across tasks, without retraining.

\end{abstract}

\section{Introduction}
\label{sec:intro}

Multimodal large language models (MLLMs) have recently achieved impressive progress in video understanding~\cite{bai2025qwen25vltechnicalreport,wang2025internvl3}.
However, this capability comes with a practical bottleneck: long videos contain far more frames than current models can process~\cite{wang2024internvideo2, shen2024longvu}.
Encoding every frame quickly exceeds the limited context and visual-token budgets of current architectures, making dense video processing computationally infeasible.
Therefore, long video understanding largely reduces to a problem: \textit{selecting a small set of informative frames that retrieves query-relevant visual evidence}.

\begin{figure}[t]
    \includegraphics[width=\linewidth]{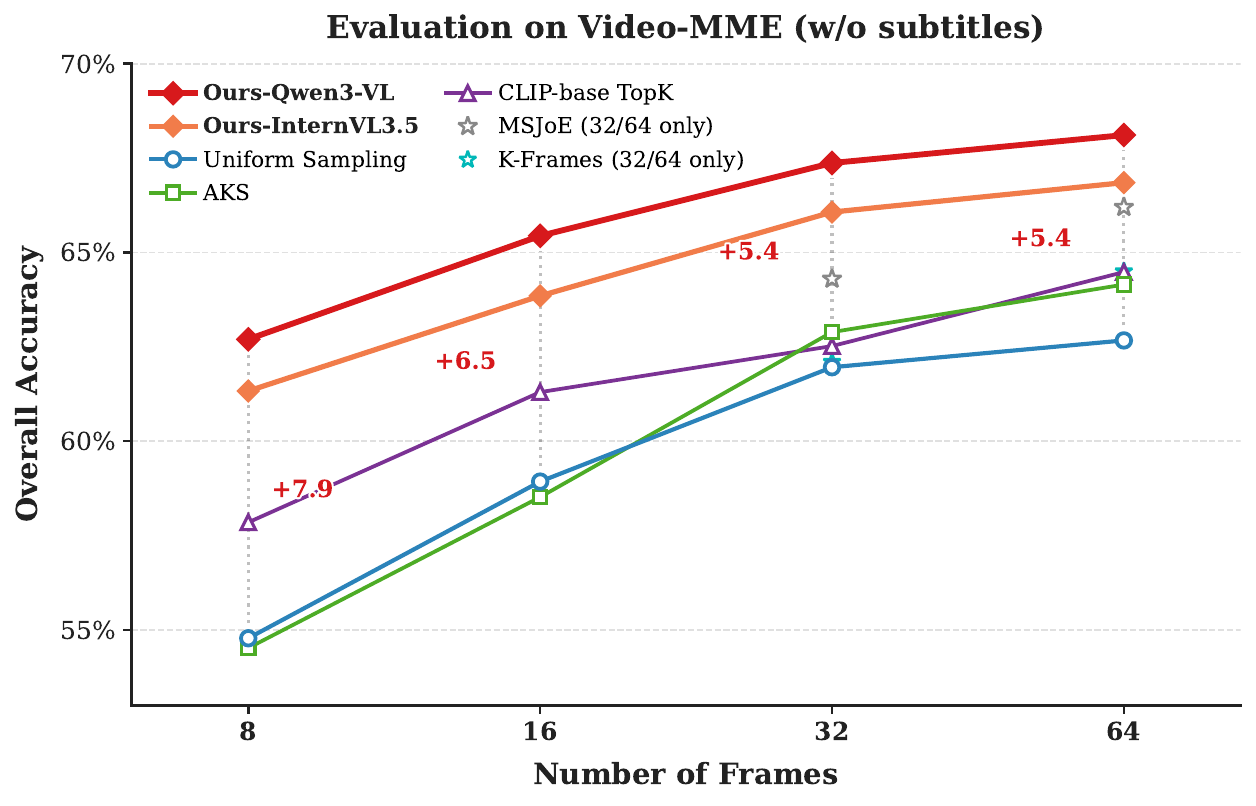}
    \caption {\textbf{Performance--efficiency overview.}
    Our selector improves long-video understanding while keeping the selection stage lightweight, illustrating the accuracy--cost trade-off targeted by this work.
    }
    \label{fig:performance_overview}
\end{figure}

\begin{figure*}[t]
    \centering
    \includegraphics[width=\textwidth,trim=12 48 12 48,clip]{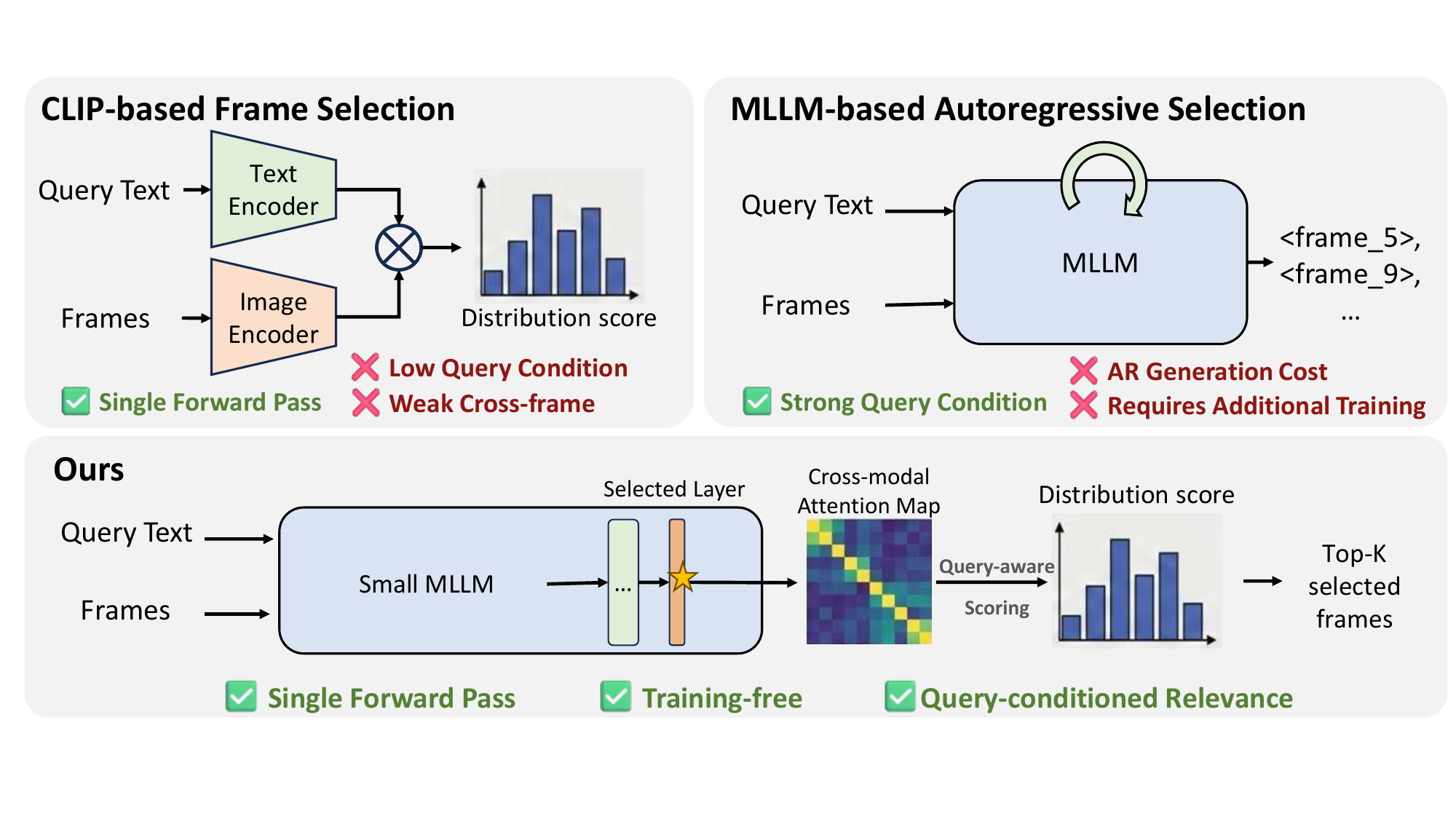}
    \caption{\textbf{Comparison of text-guided frame selection paradigms for long video understanding.}
    CLIP-based methods score frames independently with \textit{weak query conditioning}. 
    MLLM-based autoregressive selectors achieve strong query conditioning but require \textit{additional training}. 
    Our method extracts cross-modal attention from a small MLLM in a single forward pass for \textit{training-free, query-conditioned} selection.
    }
    \label{fig:overview}
\end{figure*}

Existing approaches navigate a trade-off between semantic depth and computational cost (Figure~\ref{fig:overview}).
Uniform sampling~\cite{wang2024qwen2, bai2025qwen25vltechnicalreport, lin2024video, Maaz2023VideoChatGPT} is cheap but content-blind: it treats all temporal regions as equally important and often misses sparse, decisive events in long videos.
CLIP-based methods~\cite{tang2025adaptive, zhang2025q, liu2025bolt} score frames by query--frame similarity, but do so independently, without cross-frame context or strong query conditioning.
More generally, this setting is not standard image-text retrieval.
Rather than retrieving visually similar frames in isolation, it requires selecting a compact set of complementary evidence frames that best supports long-video QA under a fixed answerer budget.
MLLM-based selectors~\cite{yao2025kframesscenedrivenanykkeyframe, xu2025viarl} offer deeper semantic reasoning.
However, feeding multiple frames into an MLLM produces a large number of visual tokens, which are constrained by the model's limited context window and thus restrict the number of frames that can be processed jointly.
In addition, the autoregressive inference paradigm introduces non-trivial latency, further increasing computational cost and limiting scalability to long video sequences.

We observe that MLLMs already expose a useful retrieval signal before answer generation.
At validation-selected extraction layers, cross-modal attention concentrates on a small set of query-relevant visual tokens and preserves frame-level evidence (Figure~\ref{fig:attention-layer-analysis}).
Rather than asking an MLLM to autoregressively select frames, we introduce \modelname (Dynamic Attention-based Budget-aware Frame Selection), which reuses this attention as a lightweight evidence score: text-to-vision attention is converted to frame-level scores through query-conditioned aggregation, enabling even a small selector to rank frames for a stronger answerer without autoregressive decoding.

The selector, however, is constrained by a finite context window, limiting both the number of candidate frames and the visual-token budget it can process.
We formulate this selector-side allocation as a \emph{budgeted discrete optimization} problem that jointly determines (i) the candidate frame pool size and (ii) the per-frame visual token allocation under a fixed context constraint, solved efficiently via dynamic programming; the final answerer still receives the selected frames in original resolution.

In summary, our contributions are threefold: (1) a lightweight sparse frame selection framework featuring a \emph{small MLLM selector} with \emph{cross-modal attention} for query-guided frame comparison in long videos; (2) a \emph{budgeted dynamic programming} formulation that jointly optimizes the candidate frame pool size and per-frame visual token allocation under context constraints; and (3) extensive experiments showing that our approach consistently outperforms uniform sampling, heuristic keyframe extraction, CLIP-based, and MLLM-based selection while remaining computationally efficient (Figure~\ref{fig:performance_overview}).

\section{Related Work}
\label{sec:related}

\subsection{MLLMs for Long-form Video}

Recent MLLMs have enabled cross-modal video reasoning~\cite{bai2025qwen25vltechnicalreport, lin2024video}.
For long-form inputs, prior work mainly improves architectures, training pipelines, visual encoders, and temporal modeling, including video-specialized MLLMs~\cite{li2025videochat,jin2024chat,cheng2024videollama}, multi-scale visual modeling~\cite{li2024llava,xu2024slowfast}, and long-context or temporally compressed designs~\cite{chen2024longvila,shu2025video,fei2024video,shen2024longvu,cheng2025scaling,team2025kwai}.
Our work is orthogonal to these efforts: we focus on test-time input construction for existing MLLMs.

\subsection{CLIP-based Frame Selection}
CLIP-based frame selection uses pretrained vision-language encoders (e.g., CLIP/SigLIP) to score query--frame similarity, often with diversity or coverage heuristics.
BOLT~\cite{liu2025bolt} uses inverse transform sampling, AKS~\cite{tang2025adaptive} balances prompt relevance and coverage under a token budget, and Q-Frame~\cite{zhang2025q} ranks frames into multiple compute tiers.
These methods are efficient but per-frame and similarity-driven, which underuse cross-frame temporal context.

\subsection{MLLM-based Frame Selection}
MLLM-based selectors can use richer query-video semantics but often incur training or inference overhead.
FFS~\cite{buch2025flexible} trains selectors with downstream losses; Frame-Voyager~\cite{yu2024frame} ranks candidates by task contribution; and ViaRL~\cite{xu2025viarl}, ReFoCUS~\cite{lee2025refocusreinforcementguidedframeoptimization}, and SeViLA~\cite{yu2023self} use reinforcement or self-learning.
GenS~\cite{yao2025generative} and Chain-of-Frames~\cite{ghazanfari2025chain} rely on keyframe supervision, while A.I.R.~\cite{zou2025air}, GIFT~\cite{ma2026gift}, and MSJoE~\cite{tan2026msjoe} use iterative reasoning, global frame utility, or coupled sampler optimization.
In contrast, we use attention from an off-the-shelf small MLLM as a lightweight selector signal while keeping the answerer fixed.

\section{Method}
\label{sec:method}

\begin{figure*}[t]
    \centering
    \includegraphics[width=\textwidth,trim=42 170 125 160,clip]{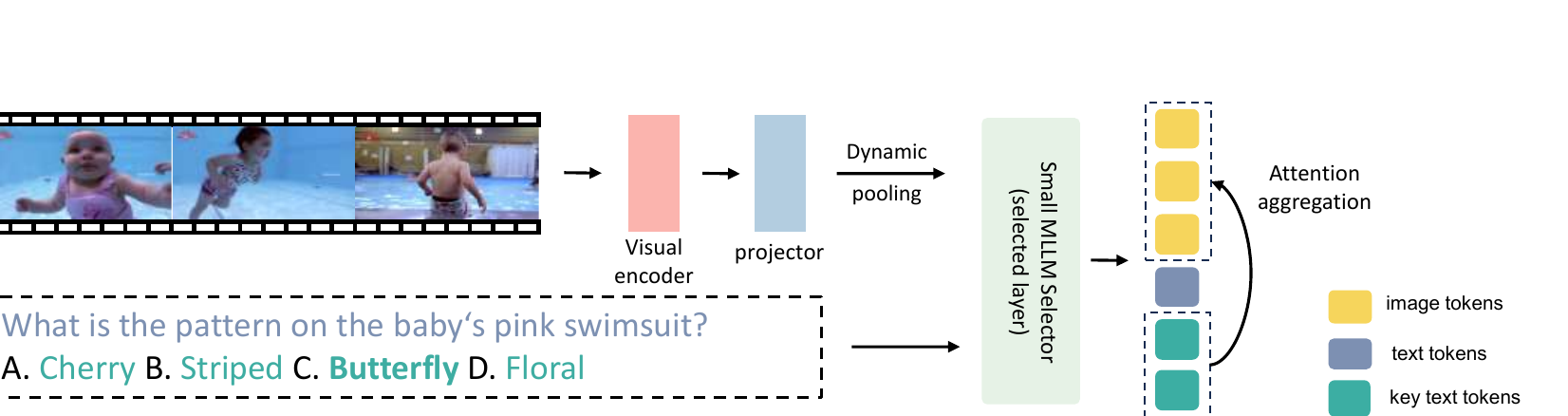}
    
    \caption{
    \textbf{Overview of \modelname.}
    Candidate frames are compressed via dynamic pooling and scored by a small MLLM using validation-selected cross-modal attention. The top-$K$ query-relevant frames are then fed to the backbone model for prediction.
    }
    \label{fig:method}
\end{figure*}

As shown in Figure~\ref{fig:method}, we propose a lightweight and budget-aware framework for retrieving query-relevant frames from long videos under a strict visual token budget.
Our approach is motivated by an empirical observation: cross-modal attention in MLLMs tends to concentrate on query-relevant visual regions.
We exploit this signal as a lightweight proxy for frame relevance, enabling efficient selection without processing the entire video with the backbone MLLM.
However, long videos typically contain far more visual tokens than the selector's input window can hold.
To ensure scalability, we introduce budget-aware frame sampling and spatial token compression to build a compact candidate representation under a controlled visual-token budget.
Finally, we formulate a dynamic programming strategy that learns an optimal duration-aware allocation policy, balancing temporal coverage and spatial detail under a fixed token budget.

\paragraph{Pipeline Overview.}
Our pipeline operates in four steps.
First, we build a candidate pool by extracting frames at a base rate of 1 FPS (adaptively increased for short videos) and uniformly subsample $F(d)$ frames by video duration $d$.
Second, we compress each sampled frame into $P(d)$ visual tokens so the total strictly fits the global budget $B$.
Third, the selector computes a query-conditioned relevance score for each candidate using text-to-vision attention over semantically important query tokens.
Finally, we rank candidates by these scores, select the top-$K$ frames for the answerer budget ($K \leq F(d)$), and feed them to the backbone MLLM.

\subsection{Attention Pattern Analysis}
\label{sec:attn_analysis}

\begin{table}[t]
  \centering
  \footnotesize
  \setlength{\tabcolsep}{3.5pt}
  \renewcommand{\arraystretch}{1.2}
  \caption{\textbf{Layer-wise grounding evidence} of the Qwen3.5-2B selector per scoring depth (metrics defined in text; higher is better).}
  \label{tab:layerwise_grounding}
  \resizebox{\linewidth}{!}{%
  \begin{tabular}{@{} l c cc ccc @{}}
    \toprule
    \multirow{2}{*}{\textbf{Layer}} & \multirow{2}{*}{\makecell{\textbf{Video}\\\textbf{-MME}}} & \multicolumn{2}{c}{\textbf{Charades-STA}} & \multicolumn{3}{c}{\textbf{NExT-GQA}} \\
    \cmidrule(lr){3-4}\cmidrule(l){5-7}
    & & \textbf{Cov@32} & \textbf{Attn} & \textbf{Hit@32} & \textbf{Cov@32} & \textbf{Attn} \\
    \midrule
    L3  & 63.6 & 0.682 & 0.352 & 0.853 & 0.252 & 0.234 \\
    L7  & 66.7 & 0.742 & 0.426 & 0.923 & 0.291 & 0.274 \\
    L11 & 67.2 & 0.768 & \textbf{0.447} & 0.933 & \textbf{0.311} & \textbf{0.289} \\
    \rowcolor{blue!6}
    L15 & \textbf{67.4} & \textbf{0.771} & 0.443 & \textbf{0.943} & 0.303 & 0.279 \\
    L19 & 66.9 & 0.745 & 0.413 & 0.900 & 0.278 & 0.264 \\
    L23 & 66.3 & 0.733 & 0.395 & 0.847 & 0.260 & 0.250 \\
    \bottomrule
  \end{tabular}
  }
\end{table}

To motivate our selector, we analyze the layer-wise cross-modal attention behavior of InternVL3.5-2B~\cite{wang2025internvl3}. 
We summarize text-to-vision attention into frame-level relevance by aggregating attention weights over the visual tokens belonging to each frame (details in Sec.~\ref{sec:frame_scoring}), ensuring a consistent notion of relevance between analysis and selection.
Concretely, we extract text-to-vision attention from a validation-selected layer $\ell^\star$ and obtain per-frame relevance scores via this aggregation.

\begin{figure*}[t]
    \centering
    \begin{minipage}{0.49\textwidth}
        \centering
        \includegraphics[width=\linewidth,trim=5 5 5 0,clip]{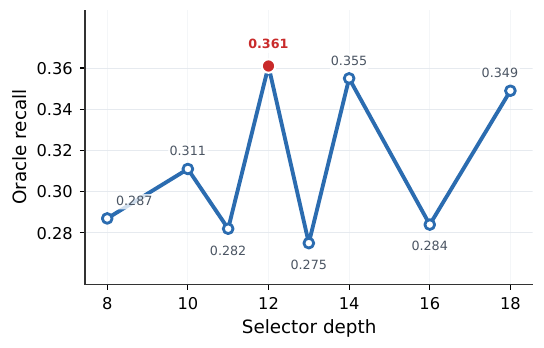}
    \end{minipage}
    \hfill
    \begin{minipage}{0.49\textwidth}
        \centering
        \includegraphics[width=\linewidth,trim=5 5 5 0,clip]{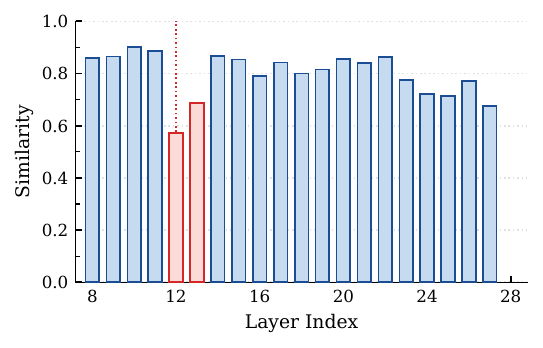}
    \end{minipage}
    \caption{\textbf{Layer-wise attention behavior.} Left: temporal-evidence recall on NExT-GQA when using each selector depth as the frame-scoring layer; depth 12 gives the strongest recall. Right: cosine similarity between the frame-level attention-score distributions of each layer and the next layer, showing how the attention pattern changes with depth.}
    \label{fig:attention-layer-analysis}
\end{figure*}

Figure~\ref{fig:attention-layer-analysis} shows the corresponding layer-wise statistics.
We find that the most effective backbone-specific layers tend to concentrate attention on a small set of query-relevant visual tokens. This pattern resembles evidence retrieval and occurs before deeper cross-modal fusion.
The NExT-GQA oracle-recall sweep in Figure~\ref{fig:attention-layer-analysis} (\emph{left}) shows that depth 12 preserves the strongest temporal evidence signal for InternVL3.5-2B, which we adopt as its extraction layer.
The \emph{right} plot measures cosine similarity between the frame-level attention-score distributions of adjacent layers.
Together, these views indicate whether a layer is both useful for selecting answer-relevant frames and distinct from neighboring layers in its attention pattern.
We observe that validation-selected layers provide the most informative relevance cues, while the similarity curve follows a decrease-then-increase trend, suggesting progressive cross-modal fusion in later layers, which obscures frame-level visual details.

To quantify this, Table~\ref{tab:layerwise_grounding} sweeps the frame-scoring depth of the Qwen3.5-2B selector on Charades-STA and NExT-GQA.
Over the top-32 scored frames, Hit@32 is the fraction of queries hitting a ground-truth frame, Cov@32 the covered fraction of the ground-truth segment, and Attn the attention mass on ground-truth frames (higher is better).
Intermediate layers (L11--L15) score highest on all three and also reach the best Video-MME accuracy, showing that mid-depth cross-modal attention carries the strongest frame-level localization signal before deeper fusion.
Our selector reuses this signal and scores frames directly from selected-layer attention, without any training.
We fix the extraction layer $\ell^\star$ using this Charades-STA/NExT-GQA sweep. Both benchmarks are disjoint from our evaluation sets (Video-MME, LVB, MLVU), so no evaluation data leaks into layer selection (Sec.~\ref{sec:exp_setup}).

\subsection{Query-aware Frame Scoring Using Selected-layer Attention}
\label{sec:frame_scoring}
We estimate frame relevance in a \textbf{query-aware} manner by converting selected-layer text-to-vision attention into frame-level scores, emphasizing informative query content during aggregation.
For brevity, we write $F=F(d)$ and $P=P(d)$ for a video of duration $d$.

\paragraph{Selected-layer Attention Scores.}
We extract attention weights from a validation-selected layer $\ell^\star$. This attention matrix represents the correlation from $T$ text tokens to all visual tokens across the sampled frames.
For multi-head attention, we average attention over heads and normalize each text-token distribution over visual tokens.
Let $A^{(\ell^\star)} \in \mathbb{R}^{T \times (F \cdot P)}$ denote the resulting text-to-vision attention matrix after mapping model-specific visual-token indices to their corresponding frames.
We compute the raw per-token relevance score for each frame by summing the attention over all visual tokens within that frame:
\begin{equation}
S_{t,j} = \sum_{p=1}^{P} A^{(\ell^\star)}_{t,(j,p)}, \quad t \in [1,T], \, j \in [1,F],
\end{equation}
where $(j,p)$ indexes the $p$-th visual token in the $j$-th frame.

\paragraph{Important-token Aggregation.}
To focus the evaluation on the core intent of the prompt, we define an important-token set $\mathcal{T}_{\mathrm{imp}}$ containing the most semantically salient tokens.
We then apply max pooling over the attention scores $S_{t,j}$ associated with these important tokens to obtain the unnormalized frame saliency $\phi_j$:
\begin{equation}
\phi_j = \max_{t \in \mathcal{T}_{\mathrm{imp}}} S_{t,j},
\qquad \phi \in \mathbb{R}^F.
\end{equation}

We specifically choose max pooling to capture the strongest relevance signal from any single key query token to the frame.
The composition of $\mathcal{T}_{\mathrm{imp}}$ is task-dependent: for multiple-choice tasks it includes the answer-option tokens; for open-ended QA we use an LLM to extract entity-centric tokens from the question.
This important-token aggregation substantially reduces positional attention bias: frequent non-informative tokens such as ``the'' often contribute noisy attention mass weakly tied to visual evidence, over-emphasizing particular temporal positions.
By scoring frames only through query-critical tokens, high-scoring frames depend more on semantic evidence than on generic attention mass.
The selector uses this attention signal in a single forward pass without autoregressive decoding or model updates.
We then rank the $F$ frames by $\phi_j$ and select the top-$K$ frames as key frames for the backbone MLLM.

\subsection{Budget-Aware Frame Sampling and Token Compression}
\label{sec:sampling_compression}
To operate under a strict context budget, we introduce two synergistic mechanisms: sparse temporal sampling and spatial token compression. The parameters for both are jointly determined by the video duration $d$ to satisfy a predefined visual token budget $B$ (i.e., the maximum context window minus the text token length).

\paragraph{Candidate Frame Pool.}

For a given video of duration $d$, we initially extract candidate frames at a base rate of 1 FPS. For short videos, we proportionally increase this rate to guarantee a comprehensive pool of at least $F(d)$ frames.
Rather than processing the entire pool, we sparsely sample $F(d)$ frames uniformly across time. The number of sampled frames $F(d)$ generally increases with $d$ to maintain adequate temporal coverage for longer videos, 
but must be balanced against the per-frame visual token allocation.

\paragraph{Token Compression.}
To fit $F(d)$ frames into the global budget $B$, each sampled frame is compressed to $P$ visual tokens \emph{after} it is encoded by the vision encoder and projected into the LLM token space.
We apply a non-overlapping $s \times s$ average pooling operation to the per-frame token grid produced by the vision encoder and projector, obtaining a reduced set of $P$ visual tokens per frame.
For efficient allocation, we select $P$ from a discrete set of perfect squares:
\begin{equation}
\mathcal{P} = \{16, 25, 36, 49, 64, 100, 144\}.
\end{equation}

\subsection{Dynamic Programming for Policy Calibration}
\label{sec:dp_policy}
Because the selector operates under a hard visual token budget $B$, allocating the optimal number of frames $F(d)$ and per-frame tokens $P(d)$ is a critical trade-off. We formulate this allocation as a discrete sequence optimization problem.

\paragraph{Monotonic Sequence Formulation.}
We discretize video durations into a set of ordered bins $d_1 < d_2 < \dots < d_N$. 
Intuitively, as video duration increases, the required number of sampled frames should not decrease (to maintain temporal coverage), and consequently, the number of tokens allocated per frame should not increase (to satisfy the budget). We enforce these as monotonicity constraints.
Our objective is to find a sequence of configurations $\{(F_i, P_i)\}_{i=1}^N$ that maximizes the sum of QA accuracy on a calibration development set, where $\mathrm{Acc}_i$ is computed by running the fixed answerer on frames selected from duration bin $i$:

\begin{equation}
\max_{\{(F_i, P_i)\}} \sum_{i=1}^N \mathrm{Acc}_i(F_i, P_i)
\end{equation}
subject to the constraints for all $i$:

\begin{align}
    F_i \cdot P_i &\le B, \\
    F_i &\le F_{i+1}, \\
    P_i &\ge P_{i+1}.
\end{align}
The three constraints enforce the context budget, non-decreasing temporal resolution, and non-increasing spatial resolution, respectively.

Because the valid choices for $(F_{i+1}, P_{i+1})$ depend on the state at step $i$, the problem exhibits optimal substructure and overlapping subproblems, so we solve for the optimal global policy with Dynamic Programming (DP) rather than exhaustive search.

\paragraph{Using the Calibrated Policy.}
During inference, an unseen video of duration $d$ is mapped to its bin $i$ and the optimal $(F_i, P_i)$ is retrieved from the DP lookup table in $O(1)$ time.
We then decode $F_i$ candidate frames, compress them to $P_i$ tokens, compute the frame saliency $\tilde{\phi}$, and select the top-$K$ frames.

Notably, the $P_i$ token compression is used only by the selector; the selected top-$K$ frames are fed to the backbone MLLM in their original uncompressed form, avoiding any degradation of the answerer input while keeping selector computation within the context limit.

\section{Experiments}
\label{sec:experiments}

\subsection{Setup}
\label{sec:exp_setup}

\paragraph{Benchmarks.}
We evaluate on three long-video benchmarks: Video-MME~\cite{fu2025video} without subtitle assistance, MLVU~\cite{zhou2025mlvu} with 3-minute to 2-hour videos, and LongVideoBench (LVB)~\cite{wu2024longvideobench} for long-range reasoning over tens of minutes to hours.
For open-ended VideoQA in Sec.~\ref{sec:exp_generalization}, we use MMBench-Video~\cite{fang2024mmbenchvideo}, which contains $\sim$600 YouTube videos and $\sim$2{,}000 QA pairs judged by GPT-4~\cite{achiam2023gpt}.

\paragraph{Implementation Details.}
We decouple frame selection and answering: a lightweight \emph{Selector} scores query-frame relevance from selected-layer cross-modal attention, while a fixed Qwen2.5-VL-7B \emph{Answerer} generates outputs from the selected frames.
Table~\ref{tab:main_results} applies the same scoring pipeline to InternVL3.5~\cite{wang2025internvl3}, Qwen2.5-VL~\cite{bai2025qwen25vltechnicalreport}, Qwen3-VL~\cite{qwen3vl2025}, and Qwen3.5~\cite{qwen35hf} selectors; the selected extraction layer for each selector is reported in the tables.
We compare with representative selection strategies, including uniform sampling, Top-$K$ SigLIP2~\cite{tschannen2025siglip} image-text scoring, BOLT~\cite{liu2025bolt}, AKS~\cite{tang2025adaptive}, and Q-Frame~\cite{zhang2025q}.
Controlled frame-selection comparisons use the same global frame budget (e.g., 32 frames), original aspect ratio, preprocessing, and sampling rules.
All evaluations are conducted with \texttt{lmms-eval}~\cite{lmms_eval2024} on NVIDIA A100 GPUs.

\paragraph{Validation Protocol.}
Our selector is training-free: it performs only \emph{inference-time calibration} and never updates any model weights.
No selector parameter is tuned on the evaluation benchmarks; calibration only fixes two inference-time operating points, the extraction layer $\ell^\star$ and the DP frame-count/token-compression allocation.
The extraction layer $\ell^\star$ is selected by the layer-wise temporal-grounding sweep on Charades-STA and NExT-GQA (Table~\ref{tab:layerwise_grounding} and Table~\ref{tab:selector_size}), which have no overlap with Video-MME, LVB, or MLVU; the resulting layers are frozen for all evaluations.
For the DP allocation, we use a separate held-out set of 1{,}000 out-of-domain videos from CG-Bench, NExT-QA, LongVideo-Reason, and LLaVA-Video (also disjoint from the evaluation benchmarks): we set the selector visual-token budget to $B{=}32{,}000$ and choose $(F(d),P(d))$ under $F(d)\cdot P(d)\le B$ using held-out QA accuracy averaged over $K \in \{16,32,64\}$.

\begin{table*}[t]
\centering
  \caption{\textbf{Main comparison of long-video frame selection methods.} Context rows are for reference; controlled rows share the same Qwen2.5-VL-7B answerer and a 32-frame budget. $^\ddagger$ denotes our reproduction. Parentheses denote selected attention layers; ``--'' means no decoupled selector or N/A. Video-MME is w/o subtitles; red superscripts show gains over uniform.}
\label{tab:main_results}

\scriptsize
\setlength{\tabcolsep}{4.2pt}
\renewcommand{\arraystretch}{0.98}
\newcommand{\scoreup}[2]{\makebox[2.9em][r]{\phantom{\up{#2}}}\makebox[2.4em][c]{#1}\makebox[2.9em][l]{\up{#2}}}

\resizebox{\textwidth}{!}{
\begin{tabular}{@{} l c l c c c @{}}
\toprule
\textbf{Method} & \textbf{Input} & \textbf{Selector} & \textbf{Video-MME} & \textbf{LVB} & \textbf{MLVU} \\
\midrule
\multicolumn{6}{@{}l}{\textbf{Closed-source MLLMs} \;{(\textit{uniform sampling; reported from original papers})}} \\
\midrule
GPT-4V~\cite{2023GPT4VisionSC} & -- & -- & 59.9 & -- & -- \\
GPT-4o~\cite{hurst2024gpt} & -- & -- & 71.9 & 66.7 & -- \\
Gemini-1.5-Pro~\cite{team2024gemini} & -- & -- & 75.0 & 64.0 & -- \\
Gemini-2.5-Pro~\cite{comanici2025gemini} & 32 fr. & -- & 77.2 & 64.2 & -- \\
\midrule
\multicolumn{6}{@{}l}{\textbf{Open-source MLLMs} \;{(\textit{uniform sampling or native video input})}} \\
\midrule
LongVILA~\cite{chen2024longvila} & 128 fr. & -- & 49.2 & -- & -- \\
Video-XL~\cite{shu2025video} & 128 fr. & -- & 55.5 & -- & 64.9 \\
LLaVA-OneVision~\cite{li2024llava} & -- & -- & 58.2 & 56.3 & 64.7 \\
LongVU~\cite{shen2024longvu} & 1 FPS & -- & 60.9 & -- & 65.4 \\
Keye-VL-1.5~\cite{team2025kwai} & 64 fr. & -- & 73.0 & 66.0 & -- \\
\midrule
\multicolumn{6}{@{}l}{\textbf{Qwen2.5-VL-7B based frame-selection methods} \;{(\textit{32 selected frames})}} \\
\midrule
Uniform & 32 fr. & -- & 61.1 & 57.6 & 62.6 \\
Top-$K$ CLIP Scoring & 32 fr. & CLIP-ViT-B/32 & 62.0 & 60.1 & 66.6 \\
Top-$K$ SigLIP2 Scoring & 32 fr. & SigLIP2-2B & 57.0 & 60.4 & 63.5 \\
Top-$K$ Qwen3 Embedding & 32 fr. & Qwen3-VL-Emb-2B & 63.3 & 62.0 & 73.3 \\
BOLT~\cite{liu2025bolt}$^\ddagger$ & 32 fr. & -- & 64.1 & 58.6 & 66.3 \\
Q-Frame~\cite{zhang2025q}$^\ddagger$ & 32 fr. & CLIP-ViT-B/32 & 63.9 & 62.0 & 67.3 \\
AKS~\cite{tang2025adaptive}$^\ddagger$ & 32 fr. & CLIP-ViT-B/32 & 62.8 & 58.8 & 69.1 \\
K-Frames~\cite{yao2025kframesscenedrivenanykkeyframe} & 32 fr. & Qwen2.5-VL-3B & 62.1 & 60.5 & 65.9 \\
A.I.R.~\cite{zou2025air} & 21 fr. & -- & 65.0 & 61.4 & 67.5 \\
GIFT~\cite{ma2026gift} & 32 fr. & -- & 64.4 & 61.3 & 68.2 \\
MSJoE~\cite{tan2026msjoe} & 32 fr. & -- & 64.3 & 60.1 & 69.3 \\
\midrule
\rowcolor{blue!6}
\textbf{\modelname-InternVL3.5} & 32 fr. & InternVL3.5-2B (L12) & \scoreup{66.2}{5.1} & \scoreup{60.3}{2.7} & \scoreup{70.5}{7.9} \\
\rowcolor{blue!6}
\textbf{\modelname-Qwen2.5-VL} & 32 fr. & Qwen2.5-VL-3B (L23) & \scoreup{66.0}{4.9} & \scoreup{60.5}{2.9} & \scoreup{71.2}{8.6} \\
\rowcolor{blue!6}
\textbf{\modelname-Qwen3-VL} & 32 fr. & Qwen3-VL-2B (L14) & \scoreup{\textbf{67.5}}{6.4} & \scoreup{62.1}{4.5} & \scoreup{74.0}{11.4} \\
\rowcolor{blue!6}
\textbf{\modelname-Qwen3.5} & 32 fr. & Qwen3.5-2B (L15) & \scoreup{67.4}{6.3} & \scoreup{\textbf{63.1}}{5.5} & \scoreup{\textbf{74.1}}{11.5} \\
\bottomrule
\end{tabular}
}
% \parbox{\textwidth}{\footnotesize\emph{Note:} $^\ddagger$ denotes our reproduction; unmarked prior selection results are reported from the original papers under the closest Qwen2.5-VL-based setting. Controlled claims use rows with the same Qwen2.5-VL-7B answerer and matched frame budget; A.I.R. uses 21 frames as a reference row. Parentheses after our selector names denote the validation-selected attention extraction layer. ``--'' means no decoupled selector or N/A. Video-MME is w/o subtitles; red superscripts are gains over uniform.}
\end{table*}

\subsection{General Video Benchmarks}
\label{sec:exp_main}
\vspace{-0.2em}

\paragraph{Quantitative Analysis.}
Table~\ref{tab:main_results} first evaluates frame selection under a fixed answerer, with controlled conclusions drawn only from rows sharing the same Qwen2.5-VL-7B answerer and matched frame budget.
In this setting, attention-based MLLM selectors improve over CLIP/SigLIP scoring and prior heuristic selectors across all three benchmarks.

With the representative InternVL3.5 selector, \modelname reaches 66.2 on Video-MME, 60.3 on LongVideoBench, and 70.5 on MLVU, outperforming BOLT and Top-$K$ CLIP scoring on Video-MME and MLVU.
Replacing only the selector backbone further improves performance: Qwen3-VL-2B reaches \textbf{67.5} on Video-MME, while Qwen3.5-2B obtains the best LongVideoBench and MLVU scores (\textbf{63.1} and \textbf{74.1}).
These results show that the pipeline is not tied to one selector: stronger lightweight selectors can be substituted while keeping the answerer unchanged.

\begin{figure}[H]
  \centering
  \includegraphics[width=\linewidth,trim=24 92 24 12,clip]{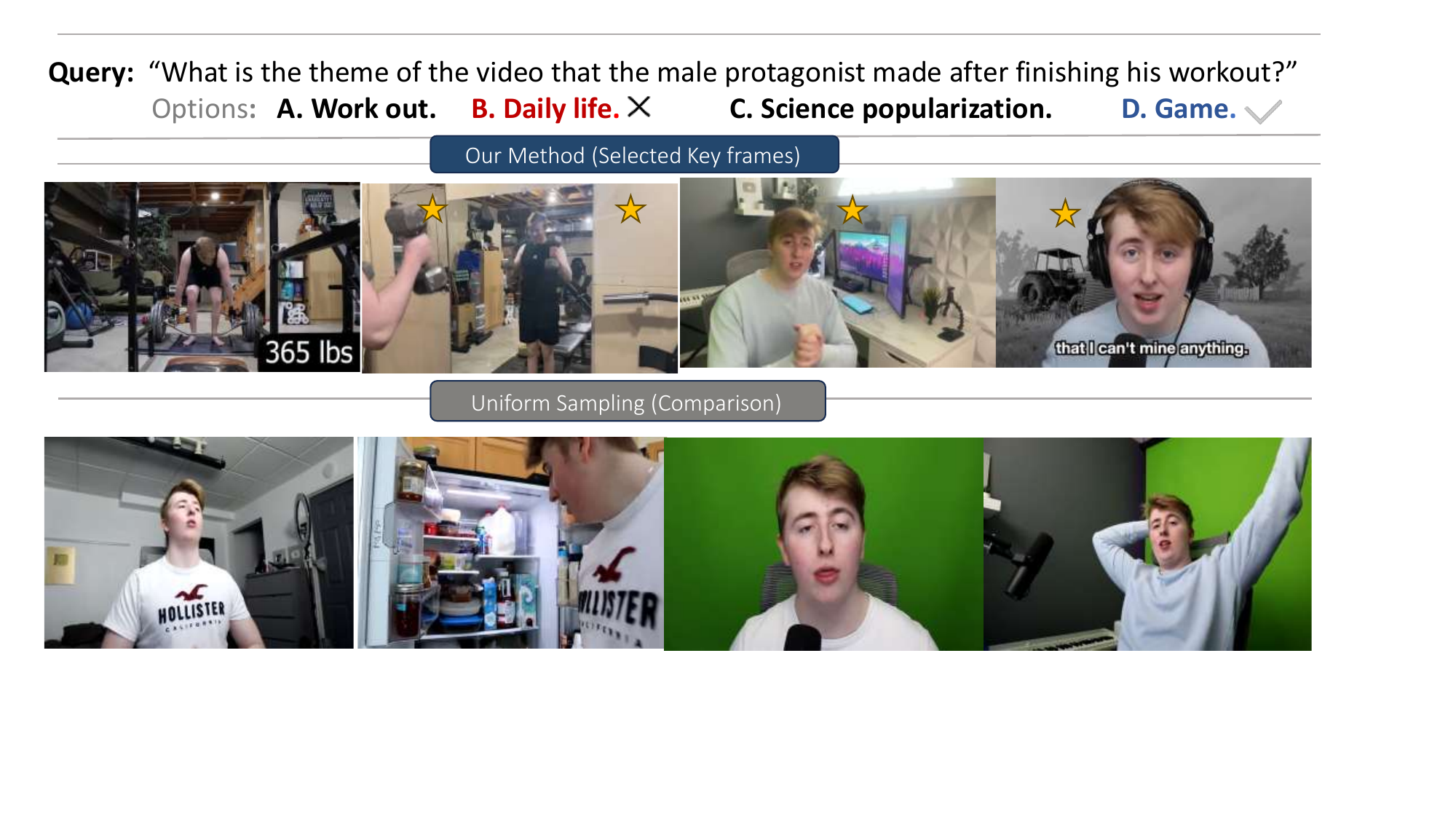}
  \caption{A Video-MME qualitative case where our selected frames provide more query-relevant evidence than uniform sampling.}
  \label{fig:case}
\end{figure}

\paragraph{Qualitative Analysis.} 
Figure~\ref{fig:case} visualizes the selected frames and corresponding answers for a representative example.
Compared with uniform sampling, our method focuses on frames that are more semantically aligned with the query, capturing both the target event and its surrounding temporal context. 
This leads to more faithful evidence for the answer and, in turn, more accurate predictions.

\paragraph{Inference Cost.}
Table~\ref{tab:end_to_end_cost} reports the full per-query wall-clock cost on Video-MME (Short), decomposed into candidate-frame decoding, selection, and answerer generation, with the answerer and frame budget fixed so that differences reflect the selection stage.
Our selection step is far cheaper than the training-based ViaRL selector ($0.47$\,s vs.\ $8.00$\,s) and more memory-efficient; even including decoding, our end-to-end latency ($6.81$\,s) stays well below ViaRL ($12.49$\,s) while reaching the best accuracy (\textbf{77.8}), showing that the accuracy gains do not come at the price of inference cost.
\begin{table}[H]
  \centering
  \footnotesize
  \setlength{\tabcolsep}{3pt}
  \renewcommand{\arraystretch}{1.2}
  \caption{\textbf{End-to-end wall-clock cost.} Per-query latency (s) on Video-MME (Short) with a fixed Qwen2.5-VL-7B answerer and $128{\to}32$ selection on a single A100.}
  \label{tab:end_to_end_cost}
  \resizebox{\linewidth}{!}{%
  \begin{tabular}{@{} l cccc cc @{}}
    \toprule
    \textbf{Method} & \textbf{Decode} & \textbf{Select} & \textbf{Answer} & \textbf{Total} & \textbf{Mem} & \textbf{Acc} \\
    \midrule
    Uniform (32) & -- & 0.00 & 1.44 & 1.44 & -- & 72.6 \\
    CLIP-ViT-B/32 & -- & 0.02 & 1.44 & -- & 0.5 & 74.0 \\
    ViaRL (Qwen2.5-VL-3B) & 3.05 & 8.00 & 1.44 & 12.49 & 7.4 & 69.7 \\
    \rowcolor{blue!6}
    \textbf{\modelname} (InternVL3.5-2B) & 4.90 & \textbf{0.47} & 1.44 & \textbf{6.81} & 3.5 & \textbf{77.8} \\
    \bottomrule
  \end{tabular}
  }
\end{table}

\paragraph{Temporal Grounding.}
We further evaluate temporal grounding on MLVU Needle-QA (N-QA), where each answer depends on a short, localized evidence segment hidden in a long video, directly testing whether the selector retrieves temporally precise evidence rather than merely covering the video uniformly.
\begin{table}[H]
  \centering
  \scriptsize
  \setlength{\tabcolsep}{1.6pt}

  \renewcommand{\arraystretch}{1.2}
  \caption{\textbf{Temporal grounding on MLVU.} All rows use Qwen2.5-VL-7B; ViaRL/K-frames use a Qwen2.5-VL-3B selector. Superscripts show gains over uniform.}
  \label{tab:viarl}
  \resizebox{\linewidth}{!}{
  \begin{tabular}{@{} l l cc cc cc @{}}
      \toprule
      \multirow{2}{*}{\textbf{Method}} & \multirow{2}{*}{\textbf{Selector}} & \multicolumn{2}{c}{\textbf{8F}} & \multicolumn{2}{c}{\textbf{16F}} & \multicolumn{2}{c}{\textbf{32F}} \\
      \cmidrule(lr){3-4}\cmidrule(lr){5-6}\cmidrule(l){7-8}
      & & \textbf{M-Avg} & \textbf{N-QA} & \textbf{M-Avg} & \textbf{N-QA} & \textbf{M-Avg} & \textbf{N-QA} \\
      \midrule
      Uniform  & -- & 57.5 & 66.7 & 61.0 & 74.5 & 62.6 & 74.9 \\
      ViaRL    & Qwen2.5-VL-3B & 58.2 & 73.5 & 61.1 & 76.1 & -- & -- \\
      K-frames & Qwen2.5-VL-3B & 60.4 & 77.5 & -- & -- & 65.9 & 79.4 \\
      \rowcolor{blue!6}
      \textbf{\modelname} & InternVL3.5-2B & \textbf{67.4}\up{9.9} & \textbf{81.1}\up{14.4} & \textbf{70.1}\up{9.1} & \textbf{81.3}\up{6.8} & \textbf{70.5}\up{7.9} & \textbf{82.5}\up{7.6} \\
      \bottomrule
  \end{tabular}
  }
  % \parbox{\linewidth}{\scriptsize\emph{Note:} All rows use Qwen2.5-VL-7B as answerer. Uniform uses no selector; ViaRL/K-frames use Qwen2.5-VL-3B. Superscripts show gains over uniform at the same frame budget.}
\end{table}

Table~\ref{tab:viarl} uses the same Qwen2.5-VL-7B answerer and matched frame budgets.
Across 8, 16, and 32 frames, our method achieves the best M-Avg and N-QA performance, consistently outperforming uniform sampling and prior training-based selectors.
The gains are largest under tighter budgets; at 8 frames, we improve over uniform sampling by \textbf{+9.9} on M-Avg and \textbf{+14.4} on N-QA.

\subsection{Generalization}
\label{sec:exp_generalization}
We test whether the selector transfers beyond the fixed-answerer setting in Table~\ref{tab:main_results}, covering both new answerers and new task formats.

\paragraph{Generalization to other models.}
Table~\ref{tab:multi-model-results} keeps the selector fixed and plugs the selected frames into frozen LLaVA-Video~\cite{zhang2024videoinstructiontuningsynthetic}, InternVL3.5-8B~\cite{wang2025internvl3}, and Qwen2.5-VL~\cite{bai2025qwen25vltechnicalreport} answerers.
Under a strict 32-frame budget, the same selector consistently improves each answerer, with gains up to \textbf{+3.5} points on Video-MME, \textbf{+4.6} on LVB, \textbf{+6.8} on MLVU M-Avg, and \textbf{+12.4} on MLVU N-QA, so the selected frames transfer beyond a particular answering backbone.

\paragraph{Generalization to other tasks.}
On MMBench-Video, we follow the official open-ended protocol with a GPT-4o-based judge.
Table~\ref{tab:mmbench_video} shows that our selector achieves the best Overall score (\textbf{1.710}), with consistent gains on most reasoning and perception categories over uniform and CLIP-based selection.

\begin{table}[t]
  % \vspace{10pt}
  \centering
  \caption{\textbf{Cross-model generalization.} The same selector is plugged into different answerers. Red superscripts show gains over uniform under a 32-frame budget.}
  \label{tab:multi-model-results}
  \scriptsize
  \setlength{\tabcolsep}{2.5pt}
  \renewcommand{\arraystretch}{1.05}
  \newcommand{\scoreplain}[1]{\makebox[2.15em][c]{#1}\makebox[2.2em][l]{}}
  \newcommand{\scoreupsmall}[2]{\makebox[2.15em][c]{#1}\makebox[2.2em][l]{\up{#2}}}
  \begin{tabular*}{\linewidth}{@{\extracolsep{\fill}} l l c c c c @{}}
    \toprule
    \multirow{2}{*}{\textbf{Answerer}} & \multirow{2}{*}{\textbf{Method}} & \multirow{2}{*}{\textbf{V-MME}} & \multirow{2}{*}{\textbf{LVB}} & \multicolumn{2}{c}{\textbf{MLVU}} \\
    \cmidrule(lr){5-6}
    & &  & & \textbf{M-Avg} & \textbf{N-QA} \\
    \midrule
    & uniform            & \scoreplain{64.0} & \scoreplain{62.9} & \scoreplain{68.6} & \scoreplain{80.0} \\
    \rowcolor{blue!6} 
    \cellcolor{white}\multirow{-2}{*}{\makecell{\textbf{InternVL3.5}\\\textbf{8B}}} & \textbf{+selector} & \scoreupsmall{\textbf{67.0}}{3.0} & \scoreupsmall{\textbf{65.8}}{2.9} & \scoreupsmall{\textbf{71.3}}{2.7} & \scoreupsmall{\textbf{82.0}}{2.0} \\
    \midrule
    & uniform            & \scoreplain{62.1} & \scoreplain{57.4} & \scoreplain{65.8} & \scoreplain{80.8} \\
    \rowcolor{blue!6} 
    \cellcolor{white}\multirow{-2}{*}{\makecell{\textbf{LLaVA-Video}\\\textbf{7B}}} & \textbf{+selector} & \scoreupsmall{\textbf{65.2}}{3.1} & \scoreupsmall{\textbf{60.1}}{2.7} & \scoreupsmall{\textbf{68.5}}{2.7} & \scoreupsmall{\textbf{85.1}}{4.3} \\
    \midrule
    & uniform            & \scoreplain{69.1} & \scoreplain{62.0} & \scoreplain{68.9} & \scoreplain{82.5} \\
    \rowcolor{blue!6} 
    \cellcolor{white}\multirow{-2}{*}{\makecell{\textbf{LLaVA-Video}\\\textbf{72B}}} & \textbf{+selector} & \scoreupsmall{\textbf{71.4}}{2.3} & \scoreupsmall{\textbf{66.6}}{4.6} & \scoreupsmall{\textbf{73.6}}{4.7} & \scoreupsmall{\textbf{85.6}}{3.1} \\
    \midrule
    & uniform            & \scoreplain{68.0} & \scoreplain{61.2} & \scoreplain{66.0} & \scoreplain{73.8} \\
    \rowcolor{blue!6} 
    \cellcolor{white}\multirow{-2}{*}{\makecell{\textbf{Qwen2.5-VL}\\\textbf{72B}}} & \textbf{+selector} & \scoreupsmall{\textbf{71.5}}{3.5} & \scoreupsmall{\textbf{65.7}}{4.5} & \scoreupsmall{\textbf{72.8}}{6.8} & \scoreupsmall{\textbf{86.2}}{12.4} \\
    \bottomrule
  \end{tabular*}
\end{table}
% \vspace{-0.5em}
\vspace{0.15em}
\begin{table}[H]
  \centering
  \scriptsize
  \setlength{\tabcolsep}{2pt}
  \renewcommand{\arraystretch}{1.1}
  
  \caption{\textbf{MMBench-Video results.} Best results are in \textbf{bold}.}
  \label{tab:mmbench_video}
  
  \resizebox{\linewidth}{!}{
  \begin{tabular}{@{} l cccc ccccc c @{}}
    \toprule
    \multirow{2}{*}{\textbf{Method}} & \multicolumn{4}{c}{\textbf{Perception}} & \multicolumn{5}{c}{\textbf{Reasoning}} & \multirow{2}{*}{\textbf{Overall}} \\
    \cmidrule(lr){2-5} \cmidrule(lr){6-10}
    & \textbf{CP} & \textbf{FP-S} & \textbf{FP-C} & \textbf{HL} & \textbf{LR} & \textbf{AR} & \textbf{RR} & \textbf{CSR} & \textbf{TR} & \\
    \midrule
    Uniform & 1.743 & 1.582 & 1.446 & 1.194 & 1.274 & 1.733 & 1.689 & 1.568 & 1.442 & 1.562 \\
    CLIP    & 1.772 & 1.721 & \textbf{1.498} & \textbf{1.226} & 1.540 & \textbf{1.808} & 1.682 & 1.543 & 1.475 & 1.657 \\
    \midrule
    \rowcolor{blue!6}
    \textbf{\modelname} & \textbf{1.852} & \textbf{1.796} & 1.465 & 1.161 & \textbf{1.611} & 1.745 & \textbf{1.818} & \textbf{1.679} & \textbf{1.512} & \textbf{1.710} \\
    \bottomrule
  \end{tabular}
  }
\end{table}
% \vspace{0.1em}

\begin{figure}[t]
    \centering
    \includegraphics[width=\linewidth]{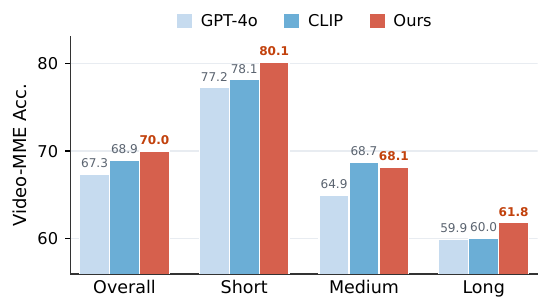}
    \caption{\textbf{Tool-use generalization.} Video-MME accuracy under the \emph{preview $\rightarrow$ retrieve $\rightarrow$ answer} workflow with GPT-4o.}
    \label{fig:tool_use}
\end{figure}

Finally, in a fixed \emph{preview $\rightarrow$ retrieve $\rightarrow$ answer} workflow (Figure~\ref{fig:tool_use}), GPT-4o inspects 16 preview frames to form a query context, then uses the retrieval module to select 16 evidence frames for answering.
Replacing CLIP retrieval with our selector improves Video-MME accuracy from $68.9\%$ to $\textbf{70.0\%}$, compared with $67.3\%$ for GPT-4o alone, so attention-based selection also serves as an external evidence-retrieval tool for a stronger model.

% \vspace{-0.4em}
% \FloatBarrier

\subsection{Ablation Study}
\label{sec:ablation}

\paragraph{Impact of Selector Size and Extraction Layer.}
Table~\ref{tab:selector_size} studies InternVL3.5 selector scale and extraction layer.
The best layer shifts with selector scale, and larger selectors bring little gain after layer calibration, showing that the selected attention layer is the main factor.

In the important-token ablation, mean pooling over $\mathcal{T}_{\mathrm{imp}}$ lowers Video-MME from \textbf{66.2} to 65.9, supporting max pooling for sparse evidence tokens.

\begin{table}[t!]
  \centering
  \scriptsize
  \setlength{\tabcolsep}{2.5pt}
  \renewcommand{\arraystretch}{1.1}
  \caption{\textbf{InternVL selector scale/layer ablation.} Qwen2.5-VL-7B is fixed. The adopted setting is highlighted.}
  \label{tab:selector_size}
  \resizebox{\linewidth}{!}{
  \begin{tabular}{@{} l c c cccc @{}}
    \toprule
    \multirow{2}{*}{\textbf{Base selector}} & \multirow{2}{*}{\begin{tabular}[c]{@{}c@{}}\textbf{Total}\\ \textbf{Layers}\end{tabular}} & \multirow{2}{*}{\textbf{Layer ($\ell^\star$)}} & \multicolumn{4}{c}{\textbf{Video-MME}} \\
    \cmidrule(l){4-7}
    & & & \textbf{Short} & \textbf{Medium} & \textbf{Long} & \textbf{Average} \\
    \midrule
    
    None (Uniform) & - & - & 72.6 & 59.0 & 51.8 & 61.1 \\
    \midrule
    
     & & L10 & 76.0 & 60.4 & 52.1 & 62.9 \\
    \rowcolor{blue!6}
    & & \textbf{L12} & \textbf{77.8} & \textbf{65.2} & \textbf{55.7} & \textbf{66.2} \\
    \multirow{-3}{*}{InternVL3.5-2B} & \multirow{-3}{*}{28} & L14 & 76.2 & 63.4 & 54.2 & 64.6 \\
    \midrule
    
     & & L12 & 73.0 & 61.9 & 51.9 & 62.2 \\
     & & \textbf{L14} & \textbf{76.4} & 65.6 & \textbf{55.3} & \textbf{65.8} \\
    \multirow{-3}{*}{InternVL3.5-4B} & \multirow{-3}{*}{32} & L16 & \textbf{76.4} & \textbf{66.3} & 52.7 & 65.2 \\
    \midrule
    
     & & L12 & 73.8 & 61.2 & 52.3 & 62.4 \\
     & & \textbf{L14} & 76.8 & \textbf{65.3} & \textbf{55.8} & \textbf{66.0} \\
    \multirow{-3}{*}{InternVL3.5-8B} & \multirow{-3}{*}{32} & L16 & \textbf{77.9} & 64.4 & 55.4 & 65.9 \\
    \bottomrule
  \end{tabular}
  }
\end{table}

\begin{table}[t!]
  \centering
  \scriptsize
  \setlength{\tabcolsep}{3pt}
  \renewcommand{\arraystretch}{1.1}
  \caption{\textbf{DP allocation ablation.} Best results are in \textbf{bold}; second-best results are \underline{underlined}.}
  \label{tab:dp_ablation}
  \resizebox{\linewidth}{!}{
  \begin{tabular}{@{} l cc cccc @{}}
    \toprule
    \multirow{2}{*}{\textbf{Strategy}} & \multirow{2}{*}{\textbf{\#Frames}} & \multirow{2}{*}{\textbf{\#Tokens}} & \multicolumn{4}{c}{\textbf{Video-MME}} \\
    \cmidrule(l){4-7}
    & & & \textbf{Short} & \textbf{Medium} & \textbf{Long} & \textbf{Average} \\
    \midrule
    \multirow{4}{*}{Static} & 64 & 144 & \underline{76.9} & 63.0 & 53.0 & 64.3 \\
     & 192 & 100 & 76.4 & \underline{64.2} & 53.3 & 64.7 \\
     & 384 & 64 & 76.4 & \textbf{65.2} & 55.3 & 65.7 \\
     & 1024 & 25 & 73.4 & 62.6 & \underline{55.6} & 63.9 \\
    \midrule
    \rowcolor{blue!6} 
    \textbf{\modelname} & \multicolumn{2}{c}{\textit{Dynamic}} & \textbf{77.8} & \textbf{65.2} & \textbf{55.7} & \textbf{66.2} \\
    \bottomrule
  \end{tabular}
  }
\end{table}

\paragraph{Effect of Dynamic Programming Allocation.}
We fix the selector and answerer and replace the duration-aware configuration $(F(d), P(d))$ with static frame/token allocations.

Table~\ref{tab:dp_ablation} shows the temporal/spatial trade-off: sparse high-token sampling helps short videos, dense low-token sampling favors long videos, and DP balances coverage with token detail to achieve the best (\textbf{66.2\%}).

\section{Conclusion}

We present \modelname, a training-free frame selector that scores query-relevant frames from selected-layer cross-modal attention of a small MLLM.
With budget-aware DP allocation, it improves reasoning across benchmarks, answerers, and tasks, and after calibration transfers to stronger answerers.

\bibliography{main}

\end{document}